\documentclass[twoside,twocolumn,9pt]{article}
\usepackage{extsizes}
\usepackage[super,sort&compress,comma]{natbib} 
\usepackage[version=3]{mhchem}
\usepackage[left=1.5cm, right=1.5cm, top=1.785cm, bottom=2.0cm]{geometry}
\usepackage{balance}
\usepackage{mathptmx}
\usepackage{sectsty}
\usepackage{graphicx} 
\usepackage{lastpage}
\usepackage[format=plain,justification=justified,singlelinecheck=false,font={stretch=1.125,small,sf},labelfont=bf,labelsep=space]{caption}
\usepackage{float}
\usepackage{fancyhdr}
\usepackage{fnpos}
\usepackage[english]{babel}
\usepackage{array}
\usepackage{droidsans}
\usepackage{charter}
\usepackage[T1]{fontenc}
\usepackage[usenames,dvipsnames]{xcolor}
\usepackage{setspace}
\usepackage[compact]{titlesec}
\usepackage[hidelinks]{hyperref}

\definecolor{cream}{RGB}{222,217,201}

\usepackage{orcidlink}
\graphicspath{{Figures/}}
\usepackage{booktabs}
\usepackage{jabbrv}

\begin{document}

\pagestyle{fancy}
\thispagestyle{plain}
\fancypagestyle{plain}{
\renewcommand{\headrulewidth}{0pt}
}

\makeFNbottom
\makeatletter
\renewcommand\LARGE{\@setfontsize\LARGE{15pt}{17}}
\renewcommand\Large{\@setfontsize\Large{12pt}{14}}
\renewcommand\large{\@setfontsize\large{10pt}{12}}
\renewcommand\footnotesize{\@setfontsize\footnotesize{7pt}{10}}
\makeatother

\renewcommand{\thefootnote}{\fnsymbol{footnote}}
\renewcommand\footnoterule{\vspace*{1pt}%
\color{cream}\hrule width 3.5in height 0.4pt \color{black}\vspace*{5pt}} 
\setcounter{secnumdepth}{5}

\makeatletter 
\renewcommand\@biblabel[1]{#1}            
\renewcommand\@makefntext[1]%
{\noindent\makebox[0pt][r]{\@thefnmark\,}#1}
\makeatother 
\renewcommand{\figurename}{\small{Fig.}~}
\sectionfont{\sffamily\Large}
\subsectionfont{\normalsize}
\subsubsectionfont{\bf}
\setstretch{1.125} 
\setlength{\skip\footins}{0.8cm}
\setlength{\footnotesep}{0.25cm}
\setlength{\jot}{10pt}
\titlespacing*{\section}{0pt}{4pt}{4pt}
\titlespacing*{\subsection}{0pt}{15pt}{1pt}

\fancyfoot{}
\fancyfoot[RO]{\footnotesize{\sffamily{\thepage}}}
\fancyfoot[LE]{\footnotesize{\sffamily{\thepage}}}
\fancyhead{}
\renewcommand{\headrulewidth}{0pt} 
\renewcommand{\footrulewidth}{0pt}
\setlength{\arrayrulewidth}{1pt}
\setlength{\columnsep}{6.5mm}
\setlength\bibsep{1pt}

\makeatletter 
\newlength{\figrulesep} 
\setlength{\figrulesep}{0.5\textfloatsep} 

\newcommand{\topfigrule}{\vspace*{-1pt}%
\noindent{\color{cream}\rule[-\figrulesep]{\columnwidth}{1.5pt}} }

\newcommand{\botfigrule}{\vspace*{-2pt}%
\noindent{\color{cream}\rule[\figrulesep]{\columnwidth}{1.5pt}} }

\newcommand{\dblfigrule}{\vspace*{-1pt}%
\noindent{\color{cream}\rule[-\figrulesep]{\textwidth}{1.5pt}} }

\makeatother

\twocolumn[
  \begin{@twocolumnfalse}
\vspace{1em}
\sffamily
\begin{center}

\LARGE{\textbf{Automating literature screening and knowledge synthesis \\ with large language models}} 
\vspace{0.4cm} 

\large{Zhanzhao Li,\orcidlink{0000-0001-7674-7424}\textit{$^{abc}$} Qiyao He,\orcidlink{0009-0005-7496-9126}\textit{$^{abc}$} Weiqiang Chen,\orcidlink{0000-0002-5355-6710}\textit{$^{abc}$} Kengran Yang,\orcidlink{0000-0001-9766-8063}\textit{$^{d}$} and Kai Gong\orcidlink{0000-0002-5884-4607}$^{\ast}$\textit{$^{abc}$}} 

\vspace{0.6cm} 

\begin{minipage}{0.9\textwidth}
\normalsize{Scientific literature is expanding rapidly, yet much of its knowledge remains difficult to organize and reuse due to the substantial workload needed to navigate large and heterogeneous literature corpora. Here, we develop a large language model (LLM)-powered framework for automating literature screening and knowledge synthesis. The framework operates on user-supplied literature corpora and integrates relevance screening, structured information extraction, synthesis, and analysis within a unified workflow. We benchmarked 11 state-of-the-art proprietary and open-weight LLMs across three materials science screening tasks differing in scope, corpus size, and class imbalance. Seven LLMs achieved balanced accuracies above 0.90 across all three tasks, with the best-performing LLM reaching 0.96--0.98, demonstrating the potential of LLMs to accurately screen large, highly imbalanced scientific corpora. The framework also substantially reduces manual effort, screening 10,000 records within four minutes compared with approximately 83 person-hours for manual screening. Beyond literature screening, the framework extracts structured information from full-text articles, enabling content-level quantification of research landscapes that extends traditional bibliometric analysis based on publication metadata. This capability reveals methodological trends and underexplored directions, as shown for machine learning in concrete materials science. We further demonstrate its utility for information and data discovery by identifying publications containing silicate glass density data that are difficult to retrieve using conventional search strategies, which can support downstream hypothesis development, research design, and data-driven scientific discovery. Overall, these results demonstrate the broad applicability of LLM-powered literature analysis across scientific domains and its potential to transform large, unstructured literature corpora into structured, analyzable knowledge at scale.} 
\end{minipage}

\end{center}

\end{@twocolumnfalse} \vspace{0.6cm}

]

\renewcommand*\rmdefault{bch}\normalfont\upshape
\rmfamily
\section*{}
\vspace{-1cm}


\footnotetext{\textit{$^{a}$~Department of Civil and Environmental Engineering, Rice University, Houston, TX 77005, USA. E-mail: kg51@rice.edu}}
\footnotetext{\textit{$^{b}$~Rice Advanced Materials Institute, Rice University, Houston, TX 77005, USA.}}
\footnotetext{\textit{$^{c}$~Ken Kennedy Institute, Rice University, Houston, TX 77005, USA.}}
\footnotetext{\textit{$^{d}$~Independent researcher, Princeton, NJ, USA.}}



\section{Introduction}

\begin{figure}[!htb]
\centering
\includegraphics{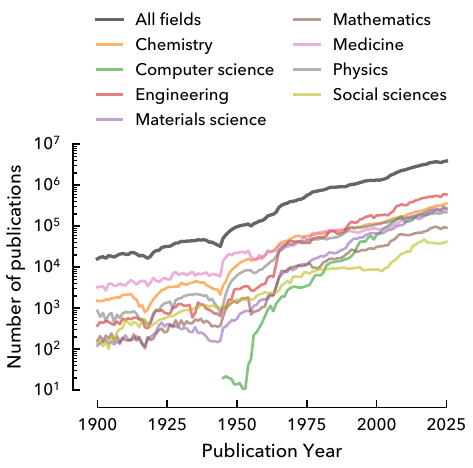}
\caption{Growth of scientific publications from 1900 to 2025 across selected disciplines. Data were obtained from Web of Science (see Table~S1 for details).}
\label{fig:introduction}
\end{figure}

Scientific literature provides the foundation for many research activities, from literature reviews and hypothesis generation to methodological design and interpretation of results. However, its rapid expansion has made it increasingly difficult for researchers to systematically navigate the literature and synthesize knowledge from it. Over the past two decades, the total number of publications has increased at an annual rate of approximately 4\%, reaching 3.9 million publications in 2025 alone (\textit{cf.}~Fig.~\ref{fig:introduction}; data from Web of Science). Despite the vast knowledge contained in this ever-expanding literature corpus, much of it remains difficult to organize and reuse systematically due to the heterogeneous scientific reporting and the substantial workload required.

Typical literature screening and knowledge synthesis workflows begin with bibliographic database searches, followed by manual review of titles and abstracts to identify potentially relevant studies, full-text inspection to determine final inclusion, and extraction of relevant information for subsequent analyses.\cite{Moher2009,Kanoulas2019} These processes are inherently labor-intensive, particularly in research areas where keyword-based searches in bibliographic databases return a large proportion of irrelevant publications for review. In some review tasks, the inclusion rate of relevant studies can be below 1\%,\cite{Dennstadt2024} requiring researchers to examine thousands of irrelevant records to identify a small number of relevant ones. A typical estimate suggests that experienced reviewers can assess approximately two title--abstract records per minute.\cite{Wallace2010} Under this assumption, screening 10,000 records would require about 83 person-hours, equivalent to more than two weeks of full-time, uninterrupted work, before any full-text review or data extraction is performed. In practice, the total time required can be substantially longer due to the complexity of the research topic, iterative refinement of research scope and inclusion criteria, independent screening by multiple reviewers, and resolution of disagreements. Human factors, such as fatigue and reduced focus during repetitive screening, together with interruptions from other research responsibilities, can further slow the process. As a result, screening may extend over weeks or months even before full-text analysis begins.\cite{Polanin2019,Joos2025} This substantial time investment can delay research progress and limit the ability to conduct timely and comprehensive reviews in rapidly evolving fields.\cite{Marshall2019} Moreover, manual workflows can be susceptible to selection bias, as one may give greater attention to highly cited papers, familiar journals, or certain research groups, which can skew the representation of the literature landscape and limit the comprehensiveness of subsequent analyses.

These challenges have motivated efforts to automate parts of the literature analysis workflow, with most existing approaches focusing on accelerating the literature screening process.\cite{VanDinter2021,Bolanos2024} Early machine learning (ML)-assisted approaches commonly used supervised classifiers, such as support vector machines, logistic regression, and na\"ive Bayes models, trained on manually labeled records to predict publication relevance.\cite{Lieng2016} These methods demonstrated the feasibility of using text classification to support screening, but they generally required topic-specific labeled datasets and had limited ability to adapt to new research questions without additional training or calibration. More recent semi-automated screening systems, including Abstrackr\cite{Wallace2012} and ASReview,\cite{VandeSchoot2021} use active learning to iteratively update the model based on reviewer feedback and prioritize records that are more likely to be relevant. Such systems can substantially reduce screening workload while maintaining high recall in many review settings. Nevertheless, they still require manually assigned relevance labels to initiate and guide screening, and their performance can be sensitive to the representativeness of those labels. Even after relevant publications are identified, downstream knowledge synthesis remains largely dependent on manual full-text interpretation, limiting large-scale and quantitative literature analyses.

Recent advances in large language models (LLMs) offer a new opportunity to address these limitations. Unlike conventional ML classifiers that typically require task-specific training data, LLMs can interpret natural-language instructions, evaluate publications against user-defined criteria, and operate in zero-shot or few-shot settings. These capabilities make them particularly promising for literature screening and knowledge synthesis. Recent studies have begun to evaluate LLMs for title and abstract screening, reporting encouraging performance and potential reductions in manual workload.\cite{Wang2024d,Li2024j,Dennstadt2024,An2024,Syriani2024,Akinseloyin2024,Jaumann2025,Joos2025} However, current applications remain limited in several important aspects. Most studies formulate the task as title- and abstract-level screening, with limited use of full-text evidence for knowledge extraction and downstream synthesis. In addition, evaluations have been concentrated largely in biomedical and clinical domains, leaving open questions about generalizability to broader scientific domains with different terminology, reporting conventions, and article structures. Commercial LLM-enabled literature discovery platforms have also emerged,\cite{Bolanos2024} but these systems often operate through platform-specific corpora and interfaces. As a result, they provide limited transparency and user control over corpus construction, screening criteria, and quantitative analysis.

Given the limitations of manual workflows and existing automation efforts, there is a clear need for frameworks that can further automate literature screening while also moving beyond relevance identification to support full-text knowledge extraction and quantitative synthesis. In this study, we develop and evaluate an LLM-powered framework for automated literature screening and knowledge synthesis. The framework operates on user-supplied literature corpora and integrates multiple stages of the review workflow, including title/abstract-based screening, full-text assessment, and structured information extraction. We benchmark 11 state-of-the-art proprietary and open-weight LLMs across three diverse materials science topics that differ in research scope, corpus size, and class imbalance, providing a realistic assessment of LLM-assisted screening under practical research scenarios. We further demonstrate the broader utility of the framework through two downstream applications of full-text analysis: quantifying the research landscape of ML applications in concrete materials science to reveal methodological trends and promising directions, and identifying publications containing silicate glass density data for database development and data-driven scientific discovery. Results show that LLMs can support not only faster literature screening, but also scalable and reproducible synthesis of scientific knowledge from large and heterogeneous literature corpora.

\section{Methodology}

\subsection{Overview}
  
\begin{figure*}[!htb]
\centering
\includegraphics{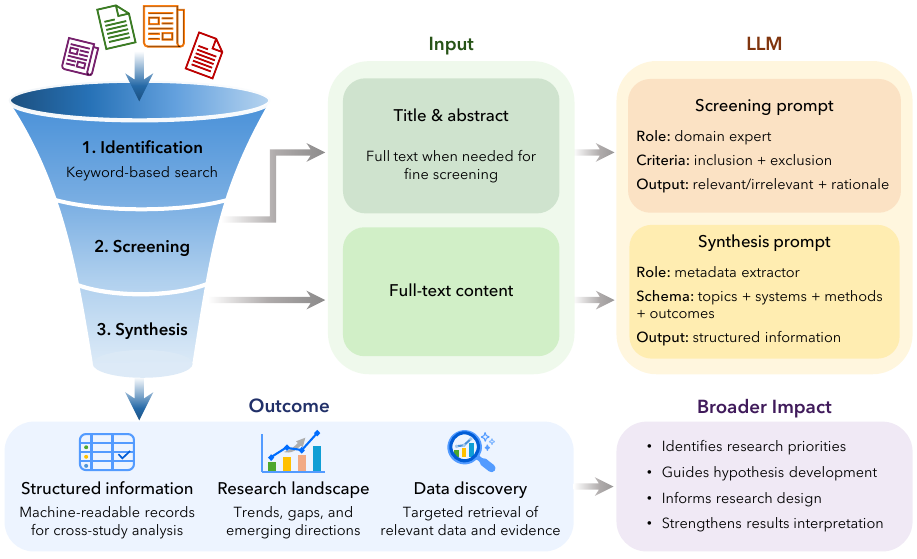}
\caption{Overview of the large language model (LLM)-powered framework for automated literature screening and knowledge synthesis.}
\label{fig:method}
\end{figure*}

The LLM-powered framework for automated literature screening and knowledge synthesis is illustrated in Fig.~\ref{fig:method}. The framework consists of three main stages: (i) identification of the initial literature corpus, (ii) screening to exclude irrelevant studies, and (iii) synthesis of knowledge from the screened literature. This study focuses on how LLMs can accelerate and enhance the latter two stages, which are typically the most time-consuming and labor-intensive components of the research workflow.

\subsubsection{Identification of initial literature corpus}

In the first stage, the research scope and objectives are defined, and an initial corpus of potentially relevant literature is retrieved using keyword-based search queries in bibliographic databases. These may include multidisciplinary databases (\textit{e.g.}, Web of Science, Scopus, and Google Scholar) as well as subject-specific databases (\textit{e.g.}, PubMed). Additional restrictions, such as publication date, document type, and language, may also be applied. The search queries are designed to be broad and inclusive to minimize the exclusion of relevant publications, with the understanding that subsequent screening stages will filter out irrelevant ones. Supplementary strategies may also be employed to improve coverage, such as checking cited or citing reference lists of key publications, searching selected journals and conference proceedings, and consulting with domain experts.\cite{Page2021a,Liberati2009} The resulting corpus is then refined by removing duplicates as well as invalid entries.

\subsubsection{Screening to exclude irrelevant studies}

Typically, the initial literature corpus is manually screened by researchers through title and abstract review to identify studies that meet predefined inclusion criteria. This process is often tedious and time-consuming, particularly when the corpus contains a large proportion of irrelevant publications and when the inclusion criteria are iteratively refined during the review process. 

The LLM-powered framework automates this screening process by leveraging the language understanding capabilities of LLMs to evaluate the relevance of each publication based on its title and abstract, and, when fine screening is required, its full text. The LLM is prompted with the research topic and predefined inclusion and exclusion criteria, and is tasked with classifying each publication as relevant or irrelevant to the research question, accompanied by a natural-language explanation summarizing the rationale for its classification (see Figs.~\ref{fig:method} and S2 for prompt examples). These explanations facilitate human inspection of sampled LLM outputs and provide a basis for iterative refinement of the screening criteria and prompt design. A publication is considered relevant only if all inclusion criteria are met and none of the exclusion criteria apply. This procedure is applied to all publications in the initial corpus, resulting in a screened set of relevant studies for downstream knowledge synthesis and analysis.

\subsubsection{Synthesis of knowledge from the screened literature}

The screened literature corpus is then analyzed to extract insights regarding the research landscape, methodological trends, and available data resources. This analysis can be performed at various levels of granularity, from high-level quantification of publication trends to detailed examination of experimental conditions and reported results.

Rather than relying on manual interpretation, which is often subjective and difficult to scale, the proposed framework leverages LLMs to extract structured information from the screened literature, including research focus, materials or systems studied, methodologies employed, and reported outcomes (see Figs.~\ref{fig:method} and S3). To enable precise extraction of relevant information that may not be fully captured in titles and abstracts, the full text of each publication is provided to the LLM at this stage. The extracted structured information is then aggregated across studies to enable systematic, quantitative, and evidence-based analysis of the literature corpus.

\subsection{Evaluation setup}

\subsubsection{Data}

To assess the performance of the LLM-powered framework and demonstrate its applicability, we apply it to diverse research topics in materials science at the intersection with data science, computational science, civil engineering, and environmental engineering. For literature screening, we select three representative topics: (i) ML in concrete materials science, (ii) molecular dynamics studies of water or ion transport in nanopores, and (iii) lithium extraction using sodium super ionic conductor (NASICON)-type materials. For knowledge synthesis, we focus on the first topic, which has a large and rapidly growing literature corpus, to showcase how the framework can be used to systematically quantify research landscapes and identify emerging research directions. Additionally, we demonstrate the capability of the framework for data discovery by identifying publications containing silicate glass density data, thereby establishing a foundation for downstream data extraction, database construction, and data-driven scientific discovery.

For each research topic, an initial literature corpus (including the title and abstract of each publication) was retrieved from Web of Science, Scopus, or Google Scholar using topic-specific search queries (\textit{cf.}~Table~S1). The retrieved corpus was then screened using the LLM-powered framework to identify relevant publications. The full texts of screened publications required for knowledge synthesis were subsequently obtained in XML format via Elsevier's ScienceDirect API, in compliance with licensing agreements.

To benchmark the LLM performance on literature screening, a subset of publications (200--440 per topic) from each of the three literature corpora was manually labeled by domain experts for relevance to the corresponding research topic. Both manual labeling and LLM screening were based on the topic-specific inclusion and exclusion criteria provided in Table~S2.

\subsubsection{LLMs}

We benchmarked 11 state-of-the-art proprietary and open-weight large language models (LLMs) for literature screening. The proprietary models included Alibaba's Qwen3.7-Max,\cite{QwenTeam2026a} Anthropic's Claude Opus 4.8,\cite{Anthropic2026} Google's Gemini 3.1 Pro,\cite{Google2026} OpenAI's GPT-5.6 Sol,\cite{OpenAI2026} and SpaceXAI's Grok 4.5.\cite{SpaceXAI2026} The open-weight models included Alibaba's Qwen3.6-35B-A3B,\cite{QwenTeam2026} DeepSeek's DeepSeek-V4-Flash,\cite{DeepSeek-AI2026} Google's Gemma 4 31B,\cite{GemmaTeam2026} Mistral AI's Mistral Medium 3.5,\cite{MistralAI2026} OpenAI's gpt-oss-120b,\cite{OpenAI2025} and Z.ai's GLM-5.2.\cite{GLM-5-Team2026} Model suffixes indicate parameter counts in billions where applicable; for Qwen3.6-35B-A3B, ``35B'' denotes 35 billion total parameters, while ``A3B'' denotes 3 billion activated parameters. These models span diverse architectures, parameter scales, training strategies, and accessibility regimes, enabling a comprehensive evaluation of LLM performance.

All models were accessed \textit{via} the OpenAI API or the OpenRouter API and were prompted using a consistent template in a zero-shot setting (see Section~S3 for details). For literature screening, models were provided with the title and abstract of each publication, whereas full-text content was used for knowledge synthesis. To reduce stochastic variation, the temperature was set to 0 whenever supported. Reasoning effort was set to the lowest supported level to evaluate the models under a computationally efficient inference configuration suitable for high-throughput literature analysis. All other configurable inference parameters were left at their API default values.

\subsubsection{Evaluation metrics}

Model performance on literature screening was evaluated using a comprehensive set of evaluation metrics, including recall (also known as sensitivity), precision, specificity, negative predictive value (NPV), $F_\beta$ scores, and balanced accuracy, as defined below:\cite{Lieng2016,Syriani2024,Almeida2016}
\begin{gather}
\text{Recall} = \frac{\mathrm{TP}}{\mathrm{TP} + \mathrm{FN}} = \frac{\text{correctly identified relevant}}{\text{all relevant}} \label{eq:recall} \\
\text{Precision} = \frac{\mathrm{TP}}{\mathrm{TP} + \mathrm{FP}} = \frac{\text{correctly identified relevant}}{\text{all identified as relevant}} \label{eq:precision} \\
\text{Specificity} = \frac{\mathrm{TN}}{\mathrm{TN} + \mathrm{FP}} = \frac{\text{correctly identified irrelevant}}{\text{all irrelevant}} \label{eq:specificity} \\
\text{NPV} = \frac{\mathrm{TN}}{\mathrm{TN} + \mathrm{FN}} = \frac{\text{correctly identified irrelevant}}{\text{all identified as irrelevant}} \label{eq:npv} \\
F_\beta = (1 + \beta^2) \cdot \frac{\text{precision} \cdot \text{recall}}{\beta^2 \cdot \text{precision} + \text{recall}} \label{eq:fbeta} \\
\text{Balanced accuracy} = \frac{\text{recall} + \text{specificity}}{2} \label{eq:balancedaccuracy}
\end{gather}
where TP, FP, TN, and FN denote the number of true positives (\textit{i.e.}, relevant publications correctly identified), false positives (irrelevant publications incorrectly identified as relevant), true negatives (irrelevant publications correctly identified), and false negatives (relevant publications incorrectly identified as irrelevant), respectively. $\beta$ is a parameter that determines the relative weight of precision and recall in the $F_\beta$ score, with $\beta > 1$ emphasizing recall and $\beta < 1$ emphasizing precision. All metrics range from 0 to 1, with higher values indicating better performance.

Recall and precision evaluate model performance for the positive class (relevant publications). Recall measures the ability to identify all relevant publications, and precision reflects the ability to include only relevant publications. Conversely, specificity and NPV evaluate the negative class (irrelevant publications). Specificity assesses the ability to exclude all irrelevant publications, and NPV measures the ability to exclude only irrelevant publications. Notably, specificity has been shown to be equivalent to normalized work saved over sampling,\cite{Kusa2023} a frequently used metric that quantifies the reduction in manual screening effort achieved by an automated screening tool compared to random sampling.\cite{Cohen2006}

To provide a more holistic assessment of model performance, we also consider the $F_1$ score ($\beta = 1$ in Eq.~\ref{eq:fbeta}), $F_2$ score ($\beta = 2$), and balanced accuracy. The $F_1$ and $F_2$ scores are weighted harmonic means of precision and recall, with the former assigning equal weight to both metrics and the latter emphasizing recall. In the context of literature screening, high recall is particularly important to minimize the risk of incorrectly excluding relevant studies; accordingly, the $F_2$ score is more appropriate.\cite{Bolanos2024} However, both $F_1$ and $F_2$ scores focus on the positive (relevant) class and do not account for performance on the negative (irrelevant) class,\cite{Derczynski2016,Powers2019a} which is also critical in literature screening. In practice, an effective screening tool should not only identify as many relevant publications as possible, but also minimize the number of irrelevant publications passed on for subsequent review, particularly because literature corpora are often highly imbalanced and dominated by irrelevant publications. To address this, we adopt balanced accuracy, which captures performance on both classes by averaging recall and specificity.

\section{Results}
\subsection{Identification of relevant literature} 
\label{sec:results_identification_of_relevant_literature}

To evaluate the performance of LLMs for automated literature screening, we applied the framework to three representative research topics: (i) ML in concrete materials science, (ii) molecular dynamics studies of water or ion transport in nanopores, and (iii) lithium extraction using NASICON-type materials. These topics were selected to represent a diverse range of research domains, from materials science to civil and environmental engineering, from experimental to computational and data-driven studies, and from broad to narrowly defined research scopes. 

For each topic, an initial corpus of publications was retrieved from Web of Science using topic-specific search queries (\textit{cf.}~Table~S1). The number of publications retrieved for each topic varied widely, with 14,595 publications for ML in concrete, 11,659 publications for nanopore transport, and 2,533 publications for lithium extraction (see Table~\ref{tab:benchmark}). This variation in corpus size reflects differences in topic scope as well as the extent of existing research activity across domains. 

\begin{table*}[!htb]
\small
\caption{Results of large language model-powered automated literature screening across three representative topics. Publication records were retrieved from Web of Science using topic-specific search queries (\textit{cf.}~Table~S1). Relevance was predicted using GPT-5.6 Sol, one of the highest-performing models evaluated during the benchmarking phase, and performance was assessed against 200--440 manually labeled publications per topic.}
\label{tab:benchmark}
\begin{tabular*}{\textwidth}{@{\extracolsep{\fill}}p{4cm}lllll}
\toprule
\textbf{Topic} 
& \textbf{Number of publications} 
& \textbf{Relevant publications (\%)} 
& \textbf{$F_1$ score ($\uparrow$)} 
& \textbf{$F_2$ score ($\uparrow$)}
& \textbf{Balanced accuracy ($\uparrow$)} \\
\midrule

Machine learning in concrete materials 
& 14,595 
& 4,515 (30.94\%) 
& 0.97
& 0.95	 
& 0.96 \\

Nanopore transport using molecular dynamics simulations
& 11,659 
& 293 (2.51\%) 
& 0.96 
& 0.97 
& 0.98 \\

Lithium extraction using NASICON-type materials 
& 2,533 
& 15 (0.59\%) 
& 0.93	 
& 0.90 
& 0.94 \\
\bottomrule
\end{tabular*}
\end{table*}

\begin{figure*}[!htb]
\centering
\includegraphics[width=\textwidth]{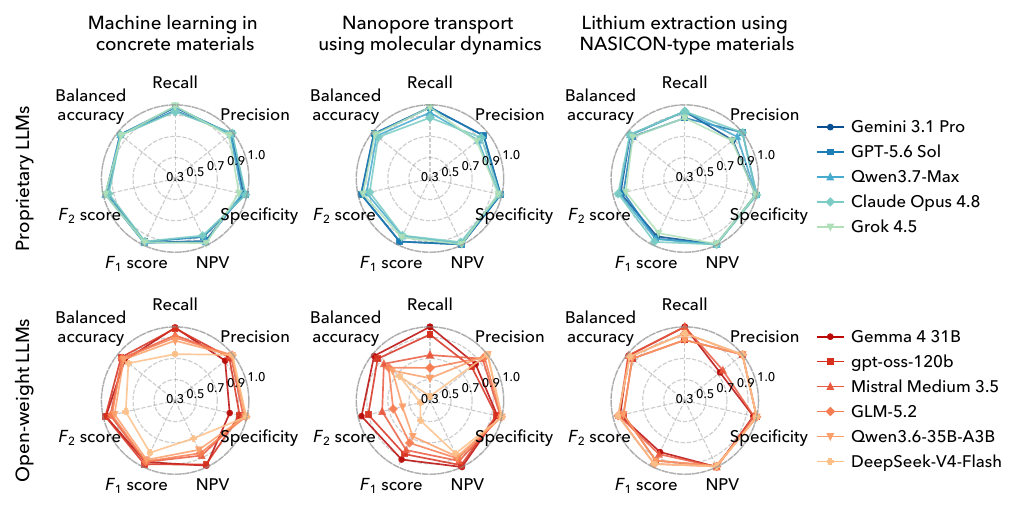}
\caption{Performance comparison of different proprietary (top row) and open-weight (bottom row) large language models (LLMs) for automated literature screening. From left to right, the panels show results for three representative topics, each evaluated against 200--440 manually labeled publications. Darker colors indicate higher mean balanced accuracy across cases, with each LLM keeping a fixed color across panels. NPV, negative predictive value.}
\label{fig:screening}
\end{figure*}

The relevance of each publication in the manually labeled datasets was then predicted based on title and abstract using a diverse set of proprietary and open-weight LLMs. Their performance is summarized in Fig.~\ref{fig:screening} and Table~\ref{tab:benchmark}. The benchmarking results demonstrate that LLMs can achieve strong zero-shot performance in literature screening. Seven of the 11 evaluated LLMs---including all five proprietary models and two of the six open-weight models---achieved balanced accuracies above 0.90 on every topic. Based on mean balanced accuracy across the three topics, the best-performing LLM achieved a mean of 0.97, with topic-specific values ranging from 0.96 to 0.98. Specifically, the proprietary LLMs performed consistently across the seven evaluation metrics, with balanced accuracies ranging from 0.93 to 0.98 and both $F_1$ and $F_2$ scores from 0.88 to 0.98. For open-weight LLMs, although the performance was more variable across topics and models, the majority of them still performed effectively. Among the 18 evaluations of the six open-weight LLMs across the three topics, balanced accuracy exceeded 0.90 in 13 evaluations, while $F_1$ and $F_2$ scores exceeded 0.90 in 10 and 9 evaluations, respectively.

Beyond screening accuracy, LLMs can substantially reduce screening time at a relatively low cost. For one of the highest-performing models, GPT-5.6 Sol, each title--abstract record required an average API processing time of 2.55 s and an average API cost of approximately \$0.0041. With a continuously replenished pool capped at 120 concurrent requests, these observed averages correspond to approximately 3.5 min and \$41 to screen 10,000 records, assuming similar input lengths and no API throttling or connection failures. Further cost reductions may be possible by using lower-cost proprietary LLMs or self-hosted open-weight LLMs with adequate performance. By comparison, at the previously reported manual screening rate of approximately two title--abstract records per minute,\cite{Wallace2010} screening 10,000 records would require about 83 person-hours, corresponding to an estimated labor cost of \$1,670 at an hourly rate of \$20. LLM-assisted screening can therefore reduce the initial screening stage from weeks of human effort to minutes of processing time and significantly lower labor costs.

Following the benchmarking phase, we deployed GPT-5.6 Sol to screen the full literature corpus for each topic (Table~\ref{tab:benchmark}). The low prevalence of relevant publications, particularly in the nanopore transport (2.51\%) and lithium extraction (0.59\%) corpora, demonstrates the potential of this approach to substantially reduce manual workload by filtering out large volumes of irrelevant records.

\subsection{Quantification of research landscape} 
\label{sec:results_quantification_of_research_landscape}

Beyond identifying relevant studies, a comprehensive, quantitative, and unbiased synthesis of these studies is essential for deriving actionable insights. Building upon the screened literature corpus for the first topic (ML in concrete materials science; 4,515 relevant publications, with full text obtained for 1,669 through the ScienceDirect API), we further demonstrate how the LLM-powered framework can be used to systematically synthesize knowledge and quantify research landscapes. Specifically, we leveraged GPT-5.6 Sol, one of the highest-performing models evaluated during the screening phase (Fig.~\ref{fig:screening} and Table~\ref{tab:benchmark}), to extract structured information from the full text of each screened publication. The information extracted includes ML tasks, algorithms employed, binder systems investigated, prediction targets, and input features used in each study (see Figs.~\ref{fig:method} and S3 for prompt examples). This structured information is then aggregated across studies to enable systematic analysis of research trends and emerging directions in this rapidly evolving field.

\begin{figure*}[!htb]
\centering
\includegraphics{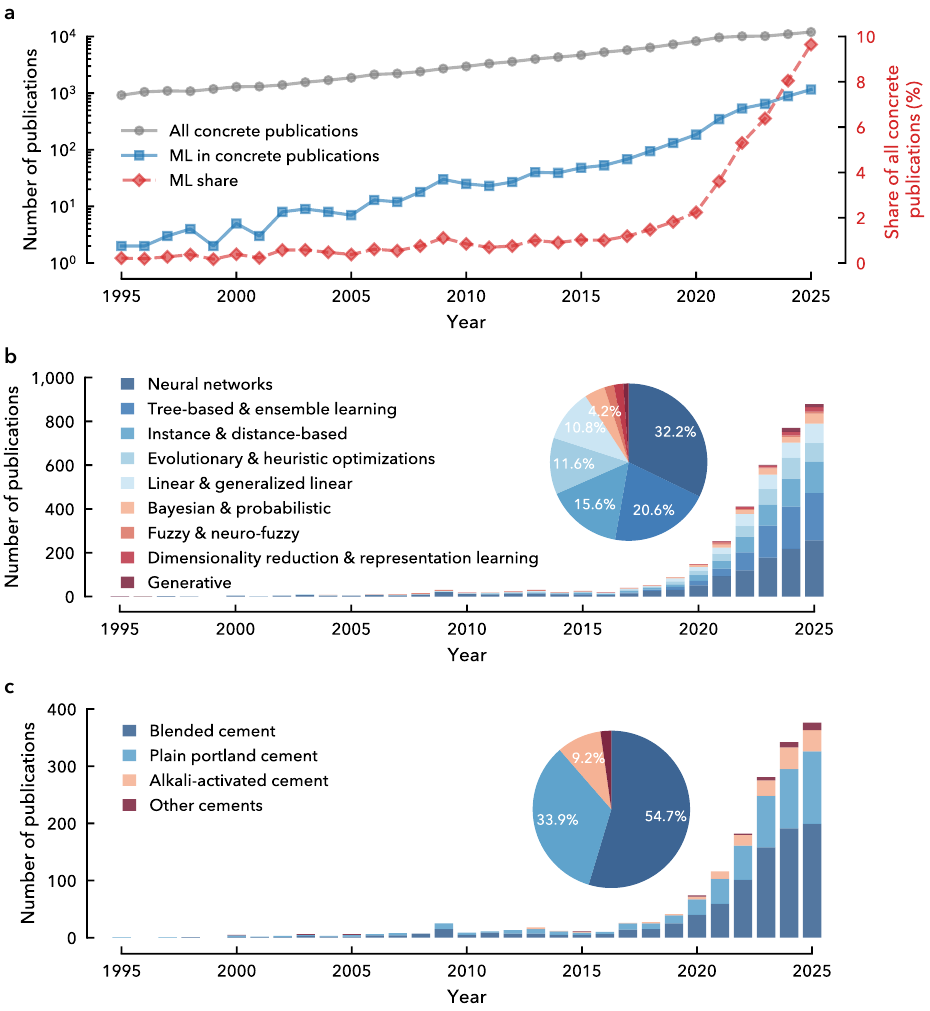}
\caption{Publication trends for machine learning (ML) applications in concrete materials science from 1995 to 2025. (a) Growth of ML-related studies compared with the broader concrete research field. (b) Distribution of ML algorithms employed. (c) Distribution of binder systems investigated. The ML-related series in panel (a) includes all 4,515 identified ML-related publications, whereas panels (b) and (c) are based on the 1,669 publications for which full text was available for analysis. Note that in (b) and (c), bars are stacked by category, with each segment representing the number of publications involving the corresponding category. As individual publications may involve multiple categories, the stacked totals for a given year do not represent the number of unique publications. Insets in (b) and (c) show the overall category distributions aggregated over all years.}
\label{fig:quantification_publication_trends}
\end{figure*}

The first level of analysis focuses on quantifying research activity in this area using the screened literature corpus. As shown in Fig.~\ref{fig:quantification_publication_trends}a, the number of publications on ML applications in concrete materials science has grown exponentially over the past three decades, with a compound annual growth rate of 23.1\% from 1995 to 2025. To provide context for this growth, we also retrieved and screened the broader literature on concrete materials science to identify all publications in this field, regardless of their use of ML. While the overall field has also experienced exponential growth, its compound annual growth rate is substantially lower at 8.9\%. This contrast highlights the rapidly increasing interest in applying ML techniques to concrete materials science, as further reflected by the growing share of ML-related studies within the broader field. Notably, the share exhibits a sharp acceleration after around 2018, increasing by approximately 6.5-fold from 1.5\% in 2018 to 9.6\% in 2025. Nearly 90\% of all identified ML-related publications were published during this period, despite the earliest studies dating back to the 1990s. This surge is likely driven by the convergence of several factors, including advances in ML methods and software ecosystems, increased availability of computational resources, and growing recognition of the potential of data-driven approaches to address complex challenges in concrete materials science.

Complementing the overall publication trends, the extracted structured information enables deeper, fine-grained analysis of methodological and topical patterns that are not readily accessible through conventional metadata-based approaches\cite{VanEck2013,Chen2014,VanEck2014,Li2022} and would otherwise require manual full-text review. As demonstrated in Fig.~\ref{fig:quantification_publication_trends}b and c, the LLM-powered framework enables systematic quantification of the distribution of ML algorithms employed in the literature as well as the binder systems investigated.

Neural networks represent the most widely used class of algorithms, accounting for the largest share of ML algorithm occurrences (32.2\%), followed by tree-based and ensemble learning methods (20.6\%) (Fig.~\ref{fig:quantification_publication_trends}b). Despite the overall dominance of neural networks, the temporal trends reveal that their share of total occurrences has declined in recent years, while that of other algorithm families---particularly tree-based and ensemble methods---has increased. This trend may reflect a growing recognition of the strengths and limitations of different algorithm families for specific research questions and data characteristics, as well as the increasing availability of user-friendly software implementations that facilitate the adoption of a broader range of methods. The pronounced growth in the use of tree-based and ensemble methods is also consistent with recent studies highlighting their strong performance on small datasets,\cite{Grinsztajn2022,Hollmann2025,Salinas2024} which are prevalent in concrete materials science.\cite{Li2026,Li2022a} Nevertheless, certain algorithm families such as Bayesian and probabilistic methods receive significantly less attention. Their potential to support uncertainty quantification and data-efficient learning\cite{Ke2021,Pfeiffer2024} suggests valuable directions for future research. Moreover, generative models, although currently underexplored, offer promising opportunities for data augmentation,\cite{BaRagaa2025,Marani2023} which can help alleviate data scarcity in concrete materials science.

In terms of binder systems, blended cement systems dominate the literature, representing 54.7\% of reported occurrences (Fig.~\ref{fig:quantification_publication_trends}c), followed by plain portland cement (33.9\%) and, to a much lesser extent, non-portland alternative binder systems (11.4\%). The predominance of blended cement reflects both the widespread use of supplementary cementitious materials and the growing emphasis on reducing the carbon footprint associated with portland cement production.\cite{Habert2020,Scrivener2022,Snellings2023} Among the alternative binder systems, alkali-activated cements, which are widely regarded as promising low-CO$_2$ alternatives to portland-based cements,\cite{Provis2014,Provis2018,Kriven2024} account for the largest share. Although their overall representation remains relatively limited, publications on alkali-activated systems have increased modestly in recent years, indicating a gradually growing interest in applying ML to these materials. In contrast, ML applications to other low-CO$_2$ binder systems---such as belite cements, calcium sulfoaluminate cements, and magnesium-based cements---remain extremely limited, with only a few studies reported in the literature. Addressing this gap by extending ML-driven investigations to these underexplored systems could support the development of a more diverse portfolio of sustainable construction materials.

\begin{figure}[!htb]
\centering
\includegraphics{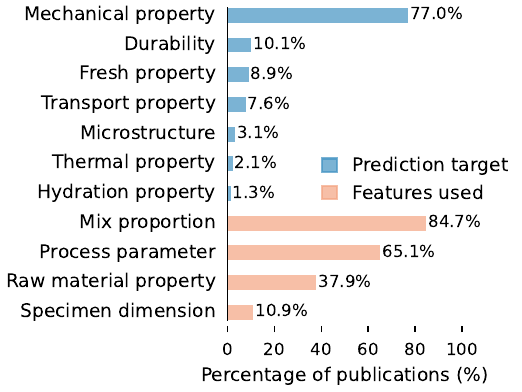}
\caption{Distribution of prediction targets and input features in machine learning studies of concrete materials, as shares of publications. Only studies involving regression tasks are considered. Note that totals exceed 100\% within each group, as individual studies can involve multiple prediction targets or input features.}
\label{fig:quantification_targets_features}
\end{figure}

Building on these method- and system-level insights, the LLM-extracted information further supports a more granular examination of what is being predicted and which information is used to support those predictions, another dimension of knowledge synthesis that is difficult to obtain without large-scale full-text analysis. As a demonstration, we narrowed the screened literature corpus to 1,334 publications involving regression tasks, which represent the most common type of ML task in this field.\cite{Li2022a} This subset of publications was then analyzed to quantify the distribution of prediction targets and input features (Fig.~\ref{fig:quantification_targets_features}). 

A clear imbalance emerges in the prediction targets investigated (Fig.~\ref{fig:quantification_targets_features}). The overwhelming majority (77.0\%) of ML studies focus on predicting mechanical properties, while other important engineering properties---such as durability (10.1\%), fresh properties (8.9\%), and transport properties (7.6\%)---are significantly underexplored. This pattern not only reflects the emphasis on mechanical performance given its central role in construction applications, but also suggests future research opportunities to broaden the scope of ML applications toward more holistic performance metrics that are essential for sustainable and resilient infrastructure. 

The analysis of input features further reveals systematic patterns in how ML models are constructed. Mix proportion variables are used in most studies (84.7\%), whereas other feature categories (\textit{e.g.}, raw material properties) are incorporated much less frequently. This imbalance may be attributed to several factors. For example, some studies are based on data generated within a single laboratory, where mix proportions are deliberately varied while other parameters are held constant and therefore not included as model inputs. In addition, the compilation of datasets for ML often prioritizes mixture design variables alongside prediction targets, whereas other potentially important descriptors are less frequently reported, curated, and thus remain less accessible for modeling.\cite{Li2026,He2026} Regardless of the underlying causes, this pattern suggests that features beyond mix proportions remain underexplored in ML studies, despite their established influence on concrete properties.\cite{Lothenbach2011,Bentz2017,Tan1996,Yi2006} As a result, models developed from such datasets may exhibit limited generalizability,\cite{Li2024} while their interpretations may overemphasize the role of mixture design and underrepresent the effects of material characteristics, processing conditions, and testing parameters. This highlights the need for more comprehensive datasets and modeling approaches that incorporate a broader range of descriptors to better capture the multivariable nature of concrete materials.

Overall, these results highlight the unique capability of the LLM-powered framework to transform large, unstructured literature corpora into structured, analyzable knowledge at scale, enabling both high-level and fine-grained insights into the research landscape. By extracting detailed information directly from full-text articles, the framework supports quantitative analyses that extend well beyond conventional bibliometric approaches based on titles, abstracts, and keywords. It enables not only the identification of overarching research trends, but also the uncovering of latent patterns, methodological biases, and underexplored directions across thousands of studies. Importantly, these analyses can be conducted in a reproducible and scalable manner, substantially reducing the time and effort required for manual synthesis.

\subsection{Discovery of information and data}

Following the quantification of the research landscape, one may narrow the focus to specific research questions and begin by collecting relevant information and data (\textit{e.g.}, existing results or methodologies) to inform hypothesis development, research design, and data-driven scientific discovery. This process requires identifying publications that contain specific types of information or extractable data, which typically involves manual interpretation of full-text articles. In this context, the LLM-powered framework can be leveraged to efficiently identify such information from large literature corpora.

Here, we demonstrate this capability by applying the framework to identify publications containing silicate glass density data, in support of downstream data extraction, database construction, and data-driven modeling. In many scientific disciplines, the availability of high-quality datasets remains a critical bottleneck to data-driven discovery.\cite{Willard2020,Cao2020,Morgan2020,Yao2023} While manually extracting data from published literature is a common approach for dataset construction, the process is typically tedious, time-consuming, and difficult to conduct exhaustively. It involves identifying potentially relevant publications from a vast literature corpus and manually interpreting full-text articles to determine whether they contain the desired data, which are often scattered across different sections and embedded in complex tables or figures\cite{Olivetti2020}---not to mention the substantial effort required for subsequent data extraction and curation. In the following, we demonstrate how the LLM-powered framework can automate and significantly accelerate these key steps.

Specifically, we leverage the framework to revisit the literature screening and data discovery process associated with the dataset developed by Gong and Olivetti,\cite{Gong2023a} in which data on chemical composition and room-temperature density of glasses in the compositional space of CaO--MgO--Al$_2$O$_3$--SiO$_2$--TiO$_2$--FeO--Fe$_2$O$_3$--Na$_2$O--K$_2$O--MnO were collected from the literature to support data-driven modeling and analysis of the composition--density relationship. Their dataset comprises approximately 2,100 data points mined from 136 studies, reflecting substantial effort in identifying and curating relevant data sources. The original literature corpus was constructed through searches in bibliographic databases, with emphasis on selected journals and disciplines, followed by manual screening and reference tracing from identified publications.

Similar to the procedures described in the earlier sections, we first retrieved an initial corpus of publications from Web of Science (\textit{cf.}~Table~S1), comprising 14,997 records. For demonstration purposes, the search was restricted to publications from Elsevier, as its API was leveraged for full-text access. To ensure a fair comparison, the publication date range was limited to studies published up to 2021, consistent with the cutoff used in the dataset of Gong and Olivetti.\cite{Gong2023a} We then applied the LLM-powered framework to perform coarse screening based on titles and abstracts to identify publications within the relevant scope (\textit{i.e.}, silicate glass), followed by fine screening based on full-text analysis to determine whether the publications contain extractable data for this task (see Table~S3 for details). In addition, the LLM was prompted to report the location of the data (\textit{e.g.}, specific sections, tables, or figures) to facilitate human inspection, and if needed, subsequent data extraction. 

\begin{figure*}[!htb]
\centering
\includegraphics{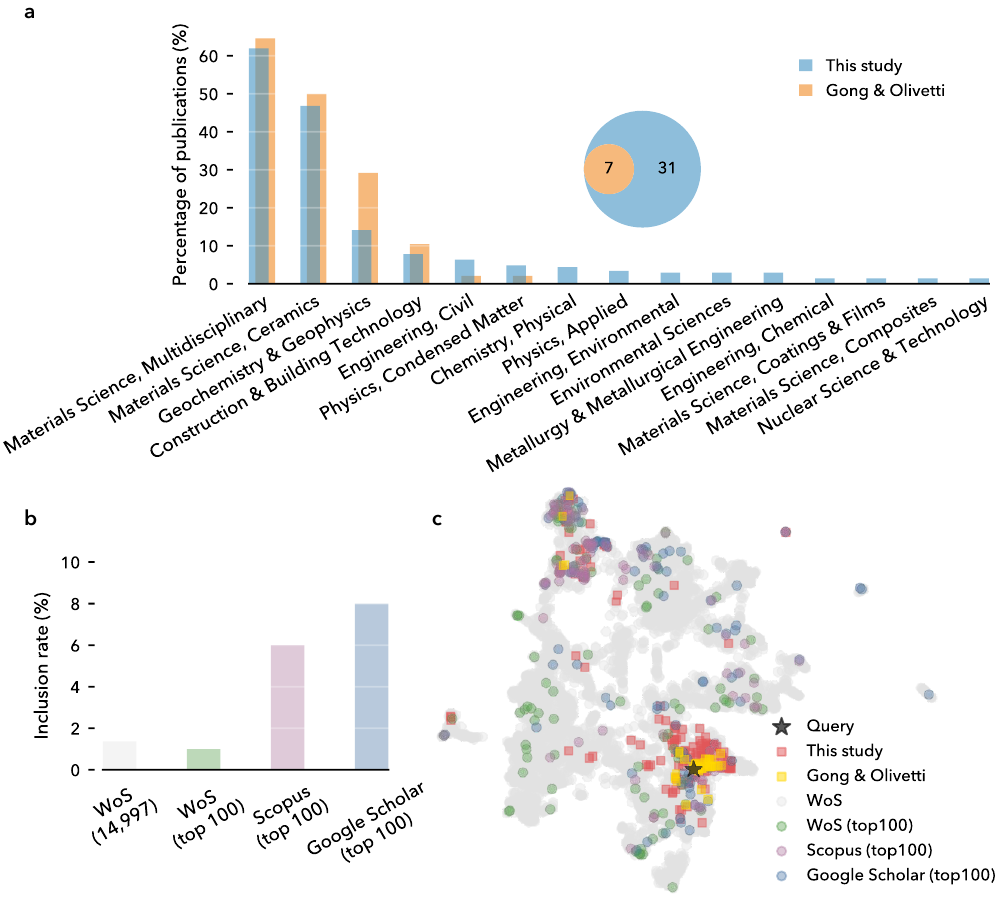}
\caption{Performance comparison of the proposed large language model (LLM)-powered framework with conventional literature retrieval methods for identifying publications containing specific information or data (here, silicate glass density data). (a) Research field coverage of relevant publications identified by the framework compared with those by Gong and Olivetti.\cite{Gong2023a} For visualization, only the top 15 fields with the highest percentages of relevant publications identified by the framework are displayed. The inset shows a Venn diagram comparing the number of research fields covered by the two approaches. (b) Inclusion rate of relevant publications across different retrieval strategies, evaluated over the full retrieved Web of Science (WoS) corpus and the top 100 search results (ranked by relevance) from WoS, Scopus, and Google Scholar, with inclusion determined using the LLM-powered framework. (c) Two-dimensional Uniform Manifold Approximation and Projection (UMAP)\cite{McInnes2020} visualization of embeddings of the query (\textit{i.e.}, the data discovery task) and retrieved publications using different methods.}
\label{fig:database}
\end{figure*}

After screening the full corpus, the LLM-powered framework identified 205 publications containing relevant data. This represents a substantial increase compared to the 136 studies included in the original dataset, and, under the same publisher restriction, 48 studies, demonstrating the improved coverage enabled by the proposed approach. In addition, the framework captures a substantially broader and more diverse range of research fields compared to those included in the original dataset (Fig.~\ref{fig:database}a; the 48 studies are considered here for a consistent comparison). Specifically, the publications identified by the framework span 38 research fields, compared to only seven fields represented in the original dataset; all seven are encompassed by the LLM-identified set (see the inset in Fig.~\ref{fig:database}a). This expanded coverage is particularly significant in light of the increasingly interdisciplinary nature of modern scientific research,\cite{VanNoorden2015,Porter2009} where information and data relevant to the research questions are often reported across diverse domains. In the context of data discovery, such breadth reduces the risk of overlooking pertinent data sources and enables the construction of more comprehensive datasets for robust and generalizable data-driven analyses.

In contrast, the narrower field coverage observed in the original dataset reflects inherent limitations of manual literature search and screening processes. To manage effort, one may be tempted to prioritize familiar journals, disciplines, or keyword spaces, which, while effective, can unintentionally bias the selection of publications. As a result, relevant studies outside these scopes---particularly those published in adjacent or emerging fields---may be overlooked, limiting the diversity and representativeness of the resulting literature corpus and, potentially, the dataset. The LLM-powered framework addresses these limitations by enabling rapid, scalable evaluation across large literature corpora, thereby identifying a more comprehensive and more exhaustive, yet less biased set of relevant publications that provides a stronger foundation for downstream research.

We further evaluate the effectiveness of conventional literature retrieval methods for identifying publications containing silicate glass density data, using Web of Science, Scopus, and Google Scholar (\textit{cf.}~Table~S1). Both the full retrieved Web of Science corpus (about 15,000 records) as well as the top 100 search results from each database (sorted by relevance and restricted here to Elsevier publications for full-text access) were screened using the LLM-powered framework (coarse followed by fine screening, as described earlier) to determine inclusion rates of relevant publications. The evaluation of the top 100 results is intended to simulate typical literature search practices, where researchers may initially focus on the highly ranked results for manual review and screening.

As shown in Fig.~\ref{fig:database}b, the inclusion rate within the full Web of Science corpus is extremely low (1.4\%), indicating that the vast majority of retrieved publications are irrelevant to the data discovery task, likely because they do not focus on silicate glass or do not report the desired data. Even when restricting the analysis to the top 100 search results, the inclusion rates remain low (1\% for Web of Science, 6\% for Scopus, and 8\% for Google Scholar), underscoring the limited precision of conventional retrieval approaches for the target task.

To further interpret these results and provide an intuitive visualization of the coverage and relevance of retrieved publications, we projected the textual embeddings of the query (\textit{i.e.}, the data discovery task) and the retrieved publications obtained by different methods into a two-dimensional space using Uniform Manifold Approximation and Projection (UMAP)\cite{McInnes2020} (Fig.~\ref{fig:database}c). The construction of these embeddings is described in Section~S4. Distances in the embedding space reflect semantic similarity, with closer proximity indicating greater similarity. In the context of data discovery, publications that are closer to the query in the embedding space are more likely to be relevant and contain the desired data. 

Publications identified by Gong and Olivetti\cite{Gong2023a} and the LLM-powered framework are densely clustered in the vicinity of the query, indicating high semantic relevance to the target task. In contrast, publications retrieved by conventional methods (\textit{i.e.}, Web of Science, Scopus, and Google Scholar) are more broadly dispersed across the embedding space, with many located far from the query. This dispersion reflects their lower relevance to the task and is consistent with the low inclusion rates observed in Fig.~\ref{fig:database}b, further highlighting the limited precision of conventional retrieval methods for identifying publications containing specific information or data.

These results highlight a fundamental challenge in using conventional retrieval methods to identify publications containing specific information or data: relevant information is often not explicitly presented in searchable fields (\textit{e.g.}, titles, abstracts, or keywords) of bibliographic databases and may be described using diverse and inconsistent terminology. Consequently, keyword-based search queries may fail to reliably capture such information. In the example of identifying publications containing silicate glass density data, enforcing highly specific keywords---such as `silicate-based glass', `room-temperature density', or specific compositional spaces---in search queries would drastically reduce the number of retrieved records, while broader queries inevitably introduce large numbers of irrelevant results and require substantial manual screening effort. 

With the LLM-powered framework, the search in bibliographic databases can instead be intentionally broad (\textit{e.g.}, using general terms such as `glass' and `density'; see Table~S1) to cover as many relevant publications as possible, with the retrieved corpus subsequently screened by LLMs at scale. This approach effectively alleviates the trade-off between recall and precision inherent in conventional retrieval methods, enabling more efficient and comprehensive discovery of information and data from the literature. The resulting literature corpus can then serve as a foundation for in-depth knowledge synthesis and downstream data extraction and analysis.

\section{Discussion}
\subsection{Lessons learned}

The results presented in this work demonstrate that the proposed LLM-powered framework can not only accelerate literature screening, but also enable deeper, more comprehensive, and more scalable synthesis of knowledge across large literature corpora. The successful applications across diverse research topics and tasks further highlight its broad applicability and versatility. As such, this framework can be integrated into many stages of scientific research workflows, from early-stage literature review and hypothesis generation to later-stage evidence synthesis and meta-analysis. To facilitate its effective deployment, we discuss several important lessons learned from the development and application of the framework.

First, the effectiveness of LLM-assisted literature screening depends strongly on how clearly the research objective is formulated and how well the inclusion and exclusion criteria are defined in the prompt. Ambiguous or poorly specified screening goals can lead to inconsistent relevance classifications. In this aspect, LLM-based screening is not fundamentally different from manual screening, which also requires clearly defined criteria that often need to be developed and refined iteratively as familiarity with the literature increases.\cite{Polanin2019} A key advantage of the LLM-powered framework, however, is that it can substantially reduce the time and effort required for each round of screening, thereby enabling rapid iteration and refinement of criteria. In addition, the LLM-generated summary of the model's rationale for each decision can be particularly valuable for identifying sources of ambiguity and guiding further prompt refinement. 

Nevertheless, defining clear criteria may not always be feasible, especially in the early stages of exploring a new research topic, when the goal is to gain a broad understanding of the landscape before more specific research questions are established. In such cases, the LLM can instead be prompted to adopt a more inclusive screening strategy and to provide explanations for its relevance judgments. In our experience, the LLM-generated explanations can also help researchers learn about a new field and identify promising directions for further investigation, particularly when the outputs of different LLMs are compared. 

Second, different LLMs may exhibit distinct decision tendencies, with some adopting a more inclusive, recall-oriented behavior and others applying more conservative criteria that improve precision at the risk of false exclusion. As shown in Fig.~\ref{fig:screening}, some open-weight LLMs exhibited higher recall than precision for lithium extraction, whereas several exhibited higher precision than recall for nanopore transport. Such differences need to be considered when selecting a model for a given task. This diversity between models can also be deliberately leveraged. For example, when the primary objective is to avoid missing relevant studies, a more inclusive model may be advantageous. By contrast, when the goal is to reduce the burden of downstream manual review, a more conservative model may be preferred. In addition, a majority voting strategy can be implemented to combine the outputs of multiple models, which may help mitigate the biases and weaknesses of individual models and improve overall screening performance.\cite{Joos2025}

Third, LLMs can generate outputs in diverse natural language expressions, which can complicate downstream quantitative analysis. In screening tasks, this challenge can be mitigated by instructing the model to return a predefined classification decision (\textit{e.g.}, relevant or irrelevant). However, for more complex tasks such as knowledge synthesis, the variability of LLM outputs can be more pronounced. For example, the same information may be reported as abbreviations or full names (\textit{e.g.}, `$k$-nearest neighbors' \textit{vs.} `KNN' for the task in Section~\ref{sec:results_quantification_of_research_landscape}), using related but non-identical terminology (\textit{e.g.}, `support vector machine' \textit{vs.} `support vector regression' in regression contexts), or at different levels of granularity (\textit{e.g.}, from broad categories like `neural network' to highly specific descriptions such as `genetic algorithm-optimized backpropagation neural network'). As a result, additional post-processing steps, such as regular expression-based parsing and manual curation, are needed to standardize the extracted information into consistent formats suitable for subsequent analysis. In this study, we addressed this challenge by prompting the LLM to report extracted information under predefined taxonomies (see Fig.~S3), which facilitated output standardization and enabled systematic analysis across studies.

\subsection{Limitations and outlook}

The capabilities of the proposed LLM-powered framework have been demonstrated on diverse research topics and use cases intentionally selected to reflect different task characteristics. Yet, these evaluations remain limited relative to the breadth of scientific literature and the diversity of real-world research workflows. Performance may vary across disciplines with different terminological conventions, article structures, or reporting practices.\cite{Joos2025} Further evaluation on a wider range of disciplines and tasks will therefore be important to establish the generalizability of the approach and to reveal domain-specific limitations as well as opportunities for improvement.

The prompt templates used in this work were developed through iterative experimentation and refinement, and they performed well across the tasks examined here. Nevertheless, they may not be optimal for all tasks or domains. We also observed that even slight changes in prompt formulation (\textit{e.g.}, screening criteria and task instructions) may lead to differences in model performance. Although a systematic investigation of prompt sensitivity is beyond the scope of this work, it represents an important direction for future research.\cite{Sclar2024,Pecher2026} Future advances in automated prompt generation and optimization\cite{Ramnath2025,Zhou2023} may further reduce reliance on manual prompt engineering while improving robustness across tasks and domains.

The current framework formulates literature screening primarily as a binary classification task, which provides a clear and practical basis for benchmarking and large-scale deployment. In many real-world review workflows, however, relevance is often better understood as a continuum rather than a strict relevant-or-irrelevant distinction (or even a three-way classification such as relevant, uncertain, or irrelevant). This allows for the possibility of relevance scoring or ranking, which could help researchers prioritize highly relevant studies among large corpora while still retaining the option to review lower-ranked studies when high recall is required, especially for interdisciplinary or exploratory topics in which inclusion boundaries are less sharply defined. Extending LLM-assisted screening beyond binary decisions to support graded relevance assessment may therefore improve both the flexibility and efficiency of the framework. Nevertheless, how the scale of such graded relevance affects model performance remains largely empirical,\cite{Zhuang2024,Jaumann2025,Dennstadt2024} and a deeper understanding will be needed to guide the design of reliable relevance scoring and ranking systems.

As noted earlier, the proposed framework operates on user-supplied literature corpora. This design helps preserve transparency, security, and reproducibility in the screening workflow, particularly in comparison with existing closed commercial platforms. At the same time, it also means that access to the underlying corpus, especially full-text content, is a prerequisite for deployment. In this work, full-text retrieval was demonstrated using Elsevier's API and XML format; yet, this framework is not restricted to these and can be easily extended to other publishers and document types (such as HTML or PDF), when access is available. In practice, such access may be constrained by publisher licensing policies and API availability, which can limit the scale and completeness of downstream analysis. Nevertheless, continued progress toward open science and more accessible publishing infrastructures will lower these barriers and create broader opportunities for large-scale automated literature analysis.

Another important limitation pertains to modality. The current pipeline focuses primarily on textual content, including paragraphs, tables, and figure captions, and does not yet fully capture information embedded directly in the visual elements of figures. This remains a significant gap, as many scientific claims, trends, and data are communicated through graphs, schematics, and images. With ongoing advances in multimodal LLMs for document understanding,\cite{Huang2025a,Alampara2025} future versions of the framework may be able to integrate textual and visual evidence more effectively, supporting more comprehensive literature analysis and evidence synthesis.

While the framework substantially accelerates and scales literature screening and synthesis by automating repetitive workloads, human oversight remains an essential component of the overall workflow. Expert input is still needed to define the scope of the analysis, refine prompts, inspect model outputs, and interpret results. The level of effort required for such oversight can vary considerably depending on task complexity, corpus diversity, and user expertise, though this investment is also intrinsic to conventional manual screening and review workflows. Moving forward, agentic AI systems with autonomy\cite{Wei2025a,Wang2024e} may reduce the need for human involvement in certain aspects of literature analysis, but careful consideration will be needed to balance automation with transparency, domain expertise, and scientific rigor.

\subsection{Broader implications}

Beyond the use cases demonstrated here, the broader significance of the LLM-powered framework lies in its potential to reshape how scientific literature is navigated, synthesized, and translated into actionable knowledge. By substantially reducing the human effort required for large-scale screening and analysis, the framework creates new opportunities for researchers to engage with the literature in more comprehensive and intellectually focused ways. Rather than being constrained by repetitive and labor-intensive review tasks, researchers can devote greater attention to interpreting patterns across studies, identifying knowledge gaps, and generating new scientific hypotheses. This shift is particularly important for research areas characterized by rapid growth and strong interdisciplinarity.

As demonstrated in this work, the framework can support not only the identification of relevant studies across broad and highly imbalanced corpora, but also the extraction of fine-grained information from full-text articles to quantify methodological choices, topical trends, and underexplored directions. More importantly, these capabilities point toward the possibility of deeper evidence-level knowledge synthesis across studies. For example, this framework could be leveraged to automatically extract and aggregate scientific claims reported in the literature, such as whether a material or treatment is reported to have positive, neutral, or negative effects on a target attribute. By quantifying the frequency and context of such observations, it may become possible to systematically analyze consensus, disagreement, and uncertainty across large bodies of literature in a more rigorous and scalable manner. Furthermore, automated extraction and synthesis of mechanistic explanations proposed for those observations could enable large-scale analysis of causal hypotheses across studies, revealing recurring explanatory patterns and highlighting areas where mechanistic understanding remains incomplete or contested. In doing so, such capabilities could support more comprehensive literature synthesis while also informing research design, guiding hypothesis generation, and strengthening interpretation of results.

These broader capabilities further lay the groundwork for a future of more automated and living literature reviews. While current applications of LLMs remain concentrated on specific components of the review process rather than full end-to-end automation,\cite{Scherbakov2025,Song2026} the present framework illustrates how literature screening, structured knowledge extraction, and subsequent synthesis can already be linked within a unified workflow. Further automation of these components, supported by user-friendly interfaces, automated prompt optimization, and tool use, will streamline the review process and further facilitate scientific research. More broadly, such capabilities align with ongoing efforts to develop AI research assistants and AI scientists that support a wider span of the research life cycle.\cite{Lu2026,Mitchener2025}

\section{Conclusions}

Literature screening and synthesis are fundamental components of scientific research, yet they have become increasingly challenging as the scientific literature continues to grow rapidly in both volume and complexity. This study demonstrates that LLMs can be leveraged to automate and accelerate these processes, enabling more comprehensive, scalable, and reproducible engagement with scientific literature. Through systematic evaluation across diverse research topics and tasks, we show that the proposed LLM-powered framework can effectively identify relevant studies from large and highly imbalanced literature corpora, support full-text quantification of research landscapes and methodological trends, and uncover data-bearing publications that are difficult to capture through conventional search strategies alone. These results highlight the potential of LLMs not merely to reduce the burden of repetitive review tasks, but to expand the scale, depth, and analytical rigor of evidence synthesis in science. As LLMs continue to advance in reasoning, multimodal understanding, tool use, and agentic workflow execution, such systems may become a core infrastructure for accelerating knowledge discovery across science and engineering.

\section*{Author contributions}

Z.L.~and K.G.~conceived this work. Z.L.~and K.Y.~developed the LLM-based framework. Z.L., Q.H., and W.C.~defined the literature screening scope for benchmarking and manually labeled paper relevance. Z.L.~performed the analyses and drafted the manuscript. Z.L.~and Q.H.~prepared the visualizations. K.G.~supervised the research and acquired funding. All authors discussed the results and contributed to the revision of the manuscript.

\section*{Conflicts of interest}
There are no conflicts to declare.

\section*{Data availability}
The three human-labeled datasets used as ground truth for evaluating literature screening performance will be made publicly available upon publication.

Supplementary information (SI): detailed bibliographic database search queries, literature screening criteria, LLM identifiers and API access details, the prompt template and examples, and the text embedding construction method. Ref.~\citenum{OpenAI2024a} is cited in the SI.

\section*{Acknowledgements}

This work was supported by the Department of Civil and Environmental Engineering at Rice University, the Rice Academy Postdoctoral Fellowship, and the Gulf Research Program's Early-Career Research Fellowship. Access to OpenAI's models was partially supported by the OpenAI Researcher Access Program. During the preparation of this work, the authors used ChatGPT (OpenAI) to assist with language refinement. The authors retained editorial control and take full responsibility for the final content. Any opinions, findings, conclusions, or recommendations expressed in this work are those of the authors and do not necessarily reflect the views of the supporting organizations.



\balance

\providecommand*{\mcitethebibliography}{\thebibliography}
\csname @ifundefined\endcsname{endmcitethebibliography}
{\let\endmcitethebibliography\endthebibliography}{}

\end{document}